\documentclass[letterpaper, 10pt, conference]{ieeeconf}  
\IEEEoverridecommandlockouts                              % This command is only needed if 
\usepackage{graphics} 
\usepackage{epsfig} 
\usepackage{mathptmx} 
\usepackage{times} 
\usepackage{amsmath} 
\usepackage{amssymb}  
\usepackage{graphicx}
\usepackage{bm}
\usepackage{eucal}

\usepackage{float}
\usepackage[ruled,vlined]{algorithm2e}
\usepackage{pgfplotstable}
\usepackage{pgfplots}
\usepackage{tikz}
\usetikzlibrary[pgfplots.groupplots]
\usepgfplotslibrary{fillbetween}
\pgfplotsset{compat=1.5}
\usepackage{subcaption}
\usepackage{caption}
\usepackage{xcolor}
\usepackage{enumerate}
\usepackage{makecell}
\usepackage{colortbl}

\usepackage{graphicx}
\usepackage{tabularx,ragged2e,booktabs}
\usepackage{url}
\usepackage{siunitx}
\usepackage{rotating}
\usepackage{amssymb}
\usepackage{tabularx}
\usepackage{pgfplots}
\usepackage{csvsimple}
\usepackage{multirow}
\usepackage{cite}
\usepackage{float}
\usepackage{graphicx}
\usepackage{booktabs}
\usepackage{todonotes}
\usepackage{multirow}
\usepackage{amsmath}
\usepackage{wrapfig}
\usepackage{cancel}
\usepackage{comment}
\usepackage{titlesec}
\usepackage[font=footnotesize]{caption}
\SetKwComment{Comment}{$\triangleright$\ }{}

\SetCommentSty{mycommfont}

\makeatletter
\renewcommand\subsubsection{\@startsection{subsubsection}{3}{\z@}%
                                     {-3.25ex\@plus -1ex \@minus -.2ex}%
                                     {-1.5ex \@plus -.2ex}% 
                                     {\normalfont\normalsize\itshape}}
\makeatother
\titlespacing*{\section}{0pt}{5pt}{1pt}
\titlespacing*{\subsection}{0pt}{2pt}{1pt}

\begin{document}
\title{Physics-Residual Dynamics and Reduced-Order Whole-Body Planning for Obstacle-Aware Human–Robot Cloth Co-Transportation}
\author{
Moein Forouhar\thanks{M. Forouhar, H. Sadeghian, and E. Steinbach are with the Technical University of Munich (TUM), Germany. E-mail: \{moein.forouhar, hamid.sadeghian, eckehard.steinbach\}@tum.de.},
Kosar Behnia\thanks{K. Behnia and G. Alcan are with Tampere University, Finland. E-mail: \{kosar.behnia, gokhan.alcan\}@tuni.fi.},
Anirvan Dutta\thanks{A. Dutta and V. Kyrki are with Aalto University, Finland. E-mail: \{anirvan.dutta, ville.kyrki\}@aalto.fi.},\\
Hamid Sadeghian,
Ville Kyrki,
Sami Haddadin\thanks{S. Haddadin is with Mohamed bin Zayed University of Artificial Intelligence (MBZUAI), UAE. E-mail: sami.haddadin@mbzuai.ac.ae.},
Gokhan Alcan,
Eckehard Steinbach
}
\maketitle

\begin{abstract}
Human--robot co-transportation of deformable objects requires predicting object deformation during motion, since obstacle clearance depends on both the grasp points and the unactuated interior. We present a hierarchical planning framework that combines a learned cloth model with a reduced-order whole-body model of a dual-arm mobile manipulator. A physics-residual conditional recurrent variational autoencoder (p-cRVAE) predicts the full cloth configuration from grasp-point observations by learning a residual correction to a computationally efficient linearized physics model, limiting error accumulation over 40-step planning horizon. The predicted cloth dynamics are embedded in a model predictive path integral (MPPI) planner using a reduced-order representation of a dual-arm mobile manipulator that preserves the non-holonomic base constraint and arm workspace limits. An MPC layer subsequently refines the sampled motion into smooth, executable references for whole-body control. The reduced-order formulation achieves tracking performance comparable to the full 17-DoF model while reducing computation time by approximately 80\%. Across four co-transportation scenarios and two carrying speeds, the proposed framework maintains cloth-obstacle clearance where a corner-following baseline results in collisions, while whole-body refinement reduces final cloth deformation from 0.93\,m to 0.28\,m.
\end{abstract}

% !TEX root = ../main.tex
\section{Introduction}
\label{sec:introduction}

Human--robot co-transportation of deformable objects is an important capability for service robotics, with applications in handling textiles, bedding, and flexible loads that are difficult for a single agent to manipulate. By sharing the task, the human and robot can distribute the support of the object while guiding it toward the desired configuration. Unlike rigid objects, deformable objects change shape under gravity and variations in their boundary conditions. Consequently, for obstacle-aware transportation, collision avoidance based only on the grasping points and initial object shape is insufficient; the evolving shape must also be considered during planning. Developing a suitable predictive model is therefore challenging due to the high-dimensional, nonlinear, and coupled dynamics of deformable objects~\cite{longhini2025unfolding}.

Physics-based models provide a physically grounded approach to capturing deformation dynamics, but their computational cost makes them impractical for motion planning. Data-driven methods offer inexpensive surrogates, but require careful strategies to achieve reliable generalization~\cite{frank2014learning}. This challenge is further complicated during execution, where only partial observations may be available. Although dense visual point clouds provide detailed surface information, occlusions and large deformations make reliable point correspondence and state estimation difficult~\cite{blanco2024benchmarking}. A practical deformable object model should therefore combine physical grounding with computational efficiency while inferring the evolving full-object configuration from sparse observations that can be reliably obtained during transportation, ideally the grasped corner points.

Beyond object modeling, co-transportation requires a motion planner that accounts for the robot model to generate feasible trajectories while maintaining the desired object geometry and avoiding environmental obstacles. Addressing these requirements in a single high-fidelity optimization problem can be computationally demanding. Sampling-based predictive methods, such as model predictive path integral (MPPI) control, offer an attractive alternative by evaluating candidate action sequences through parallel rollouts without requiring derivatives of the dynamics or cost function, and are well-suited to high-dimensional search spaces. However, finite-sample variability can produce noisy action sequences, which are undesirable when smooth reference trajectories are required. Conversely, optimization-based methods provide trajectory refinement but become increasingly challenging as the planning dimensionality and nonlinearity grow.

\begin{figure}[!t]
    \centering
    \includegraphics[width=.8\columnwidth]{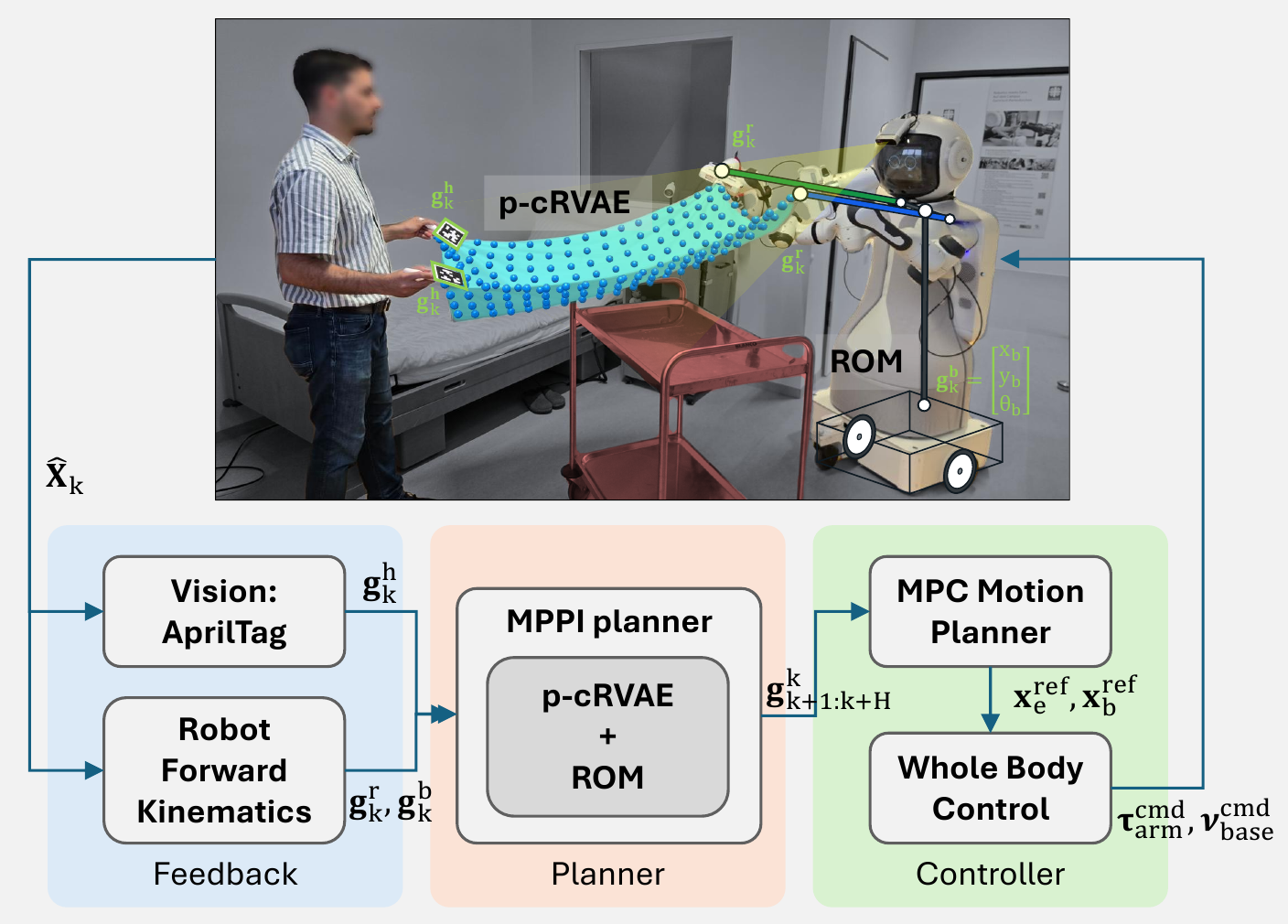}
    \caption{Setup for human–robot co-transportation of a rectangular cloth in the presence of obstacles. The planner jointly uses a high-fidelity cloth model (p-cRVAE) with a reduced-order robot model (ROM) to generate feasible hand and base trajectories while following the user and avoiding cloth–obstacle collisions.}
    \label{fig:problem_setup}
\end{figure}

Furthermore, effective co-transportation of large deformable objects benefits from a dual-arm robot with a mobile base: the two arms provide distributed support, while base mobility extends the workspace beyond the arms' stationary reach. We study this problem on GARMI~\cite{trobinger2021introducing}, which has two torque-controlled, seven-degree-of-freedom arms and a velocity-controlled, nonholonomic differential-drive base. Thus the developed planner must coordinate arm and base motions while respecting joint, workspace, and nonholonomic constraints. 

To address these challenges, we propose a hierarchical framework that combines a learned physics-residual dynamics
model with sampling- and optimization-based planning. The planning problem is deliberately divided between the
two paradigms to exploit their complementary strengths. An upper MPPI layer combines the deformable-object model
with a reduced-order robot model to efficiently explore candidate transportation motions while accounting for the
evolving cloth configuration, environmental obstacles, and robot constraints. A lower MPC layer then refines the resulting
trajectory into smooth and feasible task-space references for the robot's whole-body controller.

The key contributions of this work include:
\begin{itemize}
\item \textbf{Physics-residual cloth dynamics (p-cRVAE):} A deformable-object model combining a coarse physics-based model with a learned residual predictor to improve prediction accuracy and generalization beyond the training distribution, building on~\cite{cRVAE}.
\item \textbf{Reduced-order whole-body robot model:} A reduced-order representation of GARMI that captures the relevant arm and mobile-base workspaces and constraints, enabling efficient MPPI planning and MPC refinement.
\item \textbf{Hierarchical planning:} A sampling- and optimization-based architecture that jointly coordinates the deformable object, robot arms, and mobile base while generating smooth reference trajectories for real-time whole-body control.
\end{itemize}
We extensively evaluate the proposed framework in simulation and further assess its feasibility on the physical GARMI platform. 

\section{Related Work}
\label{sec:related_work}
Human--robot collaborative transportation of deformable objects has largely
been approached without a representation of the object's configuration.
Kruse et al.~\cite{kruse2015} keep a sheet taut through joint torque sensing
and add a depth camera that detects folds, reducing the cloth to a single
corrective direction. Nicola et al.~\cite{nicola2023} train a convolutional
network on segmented depth images to estimate the relative pose between the
robot's grasp and the operator's hands, assuming the material between the hands
to be rigid, and drive the robot to hold that pose near a rest configuration.
De Schepper et al.~\cite{deschepper2021} track the operator's skeleton
alongside force--torque measurements, inferring intention from human key points
rather than from the material. Sirintuna et al.~\cite{sirintuna2024} blend an
admittance reference with the operator's hand velocity through an adaptive
index that summarises the object's deformability in a single scalar, and apply
the same interface to a multi-robot team~\cite{sirintuna2023}. These methods do not 
use a model of the deformable object within the control framework.

A separate line of work models the configuration of deformable objects directly.
Several approaches use dense visual observations, such as point clouds or RGB
images, to encode the object into a compact latent state and predict its future
motion, as in ~\cite{li2024deformnet,ma2022graph}. These models can
capture shape details such as wrinkles and curvature, but they require visual
observations of the deformable object during inference. 
However, obtaining a continuous and unobstructed view of the
material can be difficult because of occlusions caused by the robot, the human,
or the object itself. Behnia et al.~\cite{cRVAE} avoid this requirement by
predicting the full configuration of a rope and fabric using only the measured
positions of their corner nodes. Their conditional recurrent variational
autoencoder (cRVAE) is used as the forward model of a receding-horizon controller for
obstacle-aware manipulation. 

At the robot level, whole-body control coordinates arm and base motions to track
grasp-point trajectories while managing redundancy. For safe human--robot
interaction, impedance controllers are used for both
fixed-base~\cite{sadeghian2013task} and mobile
manipulators~\cite{dietrich2016whole} to enable compliant behavior during
physical interaction with the environment. For mobile robots with nonholonomic
constraints in the base, Iskandar et al.~\cite{iskandar2019employing} first
derived the reduced-order dynamic model of the robot that is constraint-free
and then built the whole-body control framework upon it. For redundant robots,
task-priority control~\cite{sentis2006whole} organizes tasks hierarchically,
projecting lower-priority commands into the null space of higher-priority tasks
to avoid interference with their execution. Any remaining redundancy can be
exploited to accommodate additional control objectives~\cite{dietrich2011dynamic}.
Alternatively, optimization-based approaches formulate the control framework as
an optimization problem that achieves multiple task objectives through weighted
cost functions~\cite{bussmann2018whole}. In addition to task
execution, the controller must maintain the robot within its admissible
operating region. To this end, potential-based control terms have been developed
to incorporate safety-related objectives into whole-body
control~\cite{dietrich2012integration}.

\section{Problem Formulation}
\label{sec:problem}

We consider the problem of co-transporting a deformable object, such as a rectangular piece of cloth, by a human and a robot in an environment containing obstacles, while preserving the initial shape (i.e., minimizing deformation from its nominal carried configuration). The human leads the motion by holding two corners of the object, while the robot holds the opposite two and follows the human. We assume that the human trajectory is fixed over the task duration and does not account for environmental obstacles. The objective is to determine a robot grasp trajectory that maintains the desired object configuration while avoiding collisions with the environment and satisfying the robot's kinematic and workspace constraints. 

The cloth is modeled as discrete $N$ particles with positions $\mathbf{X}_k\in\mathbb{R}^{N\times3}$ at discrete time $k$. A subset $\mathcal{C}\subset\{1,\ldots,N\}$ corresponds to the grasp points and is partitioned into robot- and human-controlled particles, $\mathcal{C}=\mathcal{C}^{r}\cup\mathcal{C}^{h}$. The robot-controlled grasp points are commanded through $\mathbf{g}^{r}_k\in\mathbb{R}^{|\mathcal{C}^{r}|\times3}$, while the human grasp points $\mathbf{g}^{h}_k\in\mathbb{R}^{|\mathcal{C}^{h}|\times3}$ are measured and assumed to be approximately known over the planning horizon. Since collision avoidance cannot be based solely on the grasp points, the future cloth configuration must be predicted for candidate robot grasp trajectories.

The reference commands $\mathbf{g}^{r}_{k:k+H}$ must be tracked by the robot end-effectors at the cloth nodes $\mathcal{C}^{r}$. The robot considered in this work is GARMI, a tree-structured~\cite{hamad2023modularize} dual-arm mobile manipulator consisting of two seven-DoF arms mounted on a differential-drive mobile base. Its generalized coordinates are $\mathbf{q}_\mathrm{w}=[x,y,\theta,\mathbf{q}_{r}^{\top},\mathbf{q}_{l}^{\top}]^{\top}\in\mathbb{R}^{17}$, where $x,y,\theta$ denote the base position and orientation, and $\mathbf{q}_{r},\mathbf{q}_{l}\in\mathbb{R}^{7}$ are the joint angles of the right and left arms, respectively. The arms are actuated by ideal torque sources $\boldsymbol{\tau}_{r},\boldsymbol{\tau}_{l}\in\mathbb{R}^{7}$, while the nonholonomic mobile base is controlled by ideal velocity sources $v$ and $\omega$. The whole-body input is therefore $\boldsymbol{u}=[v,\omega,\boldsymbol{\tau}_{r}^{\top},\boldsymbol{\tau}_{l}^{\top}]^{\top}\in\mathbb{R}^{16}$, with base kinematics $\dot{x}=v\cos\theta$, $\dot{y}=v\sin\theta$, and $\dot{\theta}=\omega$, yielding the nonholonomic constraint $-\sin\theta\,\dot{x}+\cos\theta\,\dot{y}=0$. 

The robot must coordinate its arms and mobile base to track $\mathbf{g}^{r}_{k:k+H}$ while satisfying its whole-body constraints, including the admissible base workspace, arm joint limits $\mathbf{q}_{r},\mathbf{q}_{l}\in[\mathbf{q}_{\min},\mathbf{q}_{\max}]$, self-collision avoidance, and avoidance of kinematic singularities. The combined arm and base redundancy can be exploited to realize the desired grasp trajectories while satisfying these constraints. The overall problem is therefore to determine a robot grasp trajectory $\mathbf{g}^{r}_{k:k+H}$ that induces collision-free evolution of the deformable cloth while remaining feasible for the constrained robot.

\section{Methodology}
\label{sec:method}

\subsection{Cloth Model}
\label{sec:cloth}

The cloth is represented as a regular $N_u\times N_v$ particle mesh, with $N=N_uN_v$ particles having positions $\mathbf{X}_k\in\mathbb{R}^{N\times3}$ and velocities $\mathbf{V}_k$ at time $k$. The grasped particles $\mathcal{C}$ follow $\mathbf{g}_k=[\mathbf{g}^{r}_k,\mathbf{g}^{h}_k]$, while the remaining particles evolve under internal forces and gravity. We use a mesh-based Hookean mass-spring model~\cite{provot1995}, which is sufficient for the considered regime: the cloth is suspended at its four corners, remains approximately single-valued without self- or ground contact, and undergoes sufficiently slow motions such that gravitational draping dominates inertial and wrinkling effects. The model thus captures the bulk deformation relevant to obstacle clearance.

Neighboring particles are connected by structural, shear, and bending springs. With $\mathbf{d}_{ij}=\mathbf{x}_j-\mathbf{x}_i$, $\ell_{ij}=\|\mathbf{d}_{ij}\|$, and $\hat{\mathbf{u}}_{ij}=\mathbf{d}_{ij}/\ell_{ij}$, each spring obeys
\begin{equation}
  \mathbf{f}^{s}_{ij}
  =
  \left[
  k_{ij}\left(\ell_{ij}-l_{ij}\right)
  +
  c_{ij}\left((\mathbf{v}_j-\mathbf{v}_i)\cdot\hat{\mathbf{u}}_{ij}\right)
  \right]\hat{\mathbf{u}}_{ij},
  \label{eq:hooke}
\end{equation}
where $k_{ij}$, $c_{ij}$, and $l_{ij}$ are the spring stiffness, damping, and rest length. The total force on particle $i$ is
\begin{equation}
  \mathbf{f}_i
  =
  \sum_{j\in\mathcal{N}(i)}\mathbf{f}^{s}_{ij}
  -\alpha\mathbf{v}_i
  +m_i\mathbf{g}_{\mathrm{grav}}
  +\mathbf{1}[i\in\mathcal{C}]\,\mathbf{f}^{p}_i,
  \label{eq:force}
\end{equation}
where $\mathcal{N}(i)$ contains neighbors across the three spring families, $\alpha$ is the aerodynamic drag coefficient, and $m_i$ is the particle mass. Grasped particles are driven by
\begin{equation}
  \mathbf{f}^{p}_i
  =
  k_p(\mathbf{g}_{k,i}-\mathbf{x}_i)-k_d\mathbf{v}_i,
  \qquad
  k_d=2\sqrt{k_pm_i},
  \label{eq:pd}
\end{equation}
providing a critically damped response. The equations $\dot{\mathbf{v}}_i=\mathbf{f}_i/m_i$ and $\dot{\mathbf{x}}_i=\mathbf{v}_i$ are integrated over each control period $\Delta t$ using fourth-order Runge--Kutta with $n_s=100$ sub-steps and zero-order hold on $\mathbf{g}_k$. This model provides the reference dynamics for training data generation and evaluation.

\subsection{p-cRVAE Structure}
\label{sec:pcrvae}

\begin{figure}[t]
  \centering
  \includegraphics[width=.75\columnwidth]{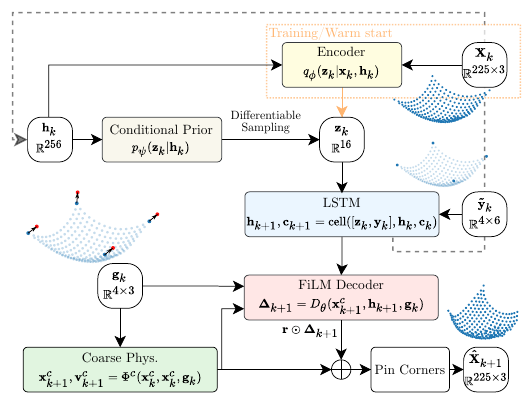}
   \caption{Structure of the proposed p-cRVAE. The coarse physics $\Phi^c$ advances the cloth state under the commanded grasp motion, and the residual decoder $D_\theta$ predicts a per-particle correction conditioned on the belief state $\mathbf{h}_k$ and the latent variable $\mathbf{z}_k$.}
  \label{fig:pcrvae}
\end{figure}

The mass-spring model in Sec.~\ref{sec:cloth} is too expensive for the sampling-based planner, requiring $\sim10^{6}$ force evaluations over the prediction horizon. We therefore leverage the conditional recurrent variational autoencoder (cRVAE) of~\cite{cRVAE} as a learned surrogate for predicting the full cloth configuration from sparse grasp-point observations. We introduce key architectural modifications and augment it with a coarse physics backbone to improve rollout accuracy and generalization (see Sec.~\ref{sec:exp}), referring to the resulting model as p-cRVAE.

Instead of directly predicting the cloth state, p-cRVAE learns a residual correction to the coarse physics prediction. As shown in Fig.~\ref{fig:pcrvae}, it consists of five components: a coarse physics backbone, recognition network, conditional prior, recurrent state, and residual decoder. The recognition network infers a stochastic latent variable from the observed state during training, while the conditional prior provides it during deployment when the full state is unavailable. The recurrent state accumulates past measurements, and the residual decoder predicts a per-particle correction to the coarse prediction.

\textbf{Coarse physics backbone.}
The coarse model advances the cloth state under the candidate grasp motion,
\begin{equation}
  \mathbf{x}^c_{k+1},\,\mathbf{v}^c_{k+1}
  =
  \Phi^c\!\left(
  \mathbf{x}^c_k,\mathbf{v}^c_k,\mathbf{g}_k
  \right)
  \label{eq:coarse}
\end{equation}
where the grasped particles are prescribed by $\mathbf{g}_k$. To make the model sufficiently inexpensive for the large number of rollouts required by MPPI, we introduce two approximations to the mass-spring cloth model. First, the grasped particles are prescribed kinematically rather than driven through the impedance-controlled gripper in \eqref{eq:pd}. Over each control period, they move linearly from their current positions to the commanded positions, making the grasp trajectory an exact boundary condition and removing the stiff gripper dynamics. Second, while Hooke's law \eqref{eq:hooke} is linear in the elongation, $\mathbf{f}^{s}$ is nonlinear in the configuration. Stacking positions and velocities as $\mathbf{x},\mathbf{v}\in\mathbb{R}^{3N}$, we therefore replace it by its first-order expansion about a gravity-settled reference  $\mathbf{x}_{ref}$\cite{baraff1998}. 

\begin{equation}
  \mathbf{f}^{s}
  \approx
  \mathbf{f}^{s}_{\mathrm{ref}}
  -\mathbf{K}(\mathbf{x}-\mathbf{x}_{\mathrm{ref}})
  -\mathbf{C}\mathbf{v},
  \label{eq:linforce}
\end{equation}
with $\mathbf{f}^{s}_{\mathrm{ref}}=\mathbf{f}^{s}(\mathbf{x}_{\mathrm{ref}},\mathbf{0})$ and constant tangent stiffness and damping
\begin{equation}
  \mathbf{K}
  =
  -\frac{\partial\mathbf{f}^{s}}{\partial\mathbf{x}}
  \bigg|_{(\mathbf{x}_{\mathrm{ref}},\mathbf{0})},
  \qquad
  \mathbf{C}
  =
  -\frac{\partial\mathbf{f}^{s}}{\partial\mathbf{v}}
  \bigg|_{(\mathbf{x}_{\mathrm{ref}},\mathbf{0})}.
  \label{eq:tangent}
\end{equation}
Substituting \eqref{eq:linforce} into $m\dot{\mathbf{v}}=\mathbf{f}^{s}-\alpha\mathbf{v}+m\mathbf{g}_{\mathrm{grav}}$ and applying backward Euler with $\mathbf{x}_{t+1}=\mathbf{x}_t+h\,\mathbf{v}_{t+1}$ leaves a single linear solve per sub-step,

\begin{equation}
  \begin{aligned}
    \mathbf{A}\mathbf{v}_{t+1}
    &=
    \frac{m}{h}\mathbf{v}_t
    +\mathbf{f}^s_{ref}
    -\mathbf{K}(\mathbf{x}_t-\mathbf{x}_{ref})
    +m\mathbf{g}_{\mathrm{grav}},\\
    \mathbf{A}
    &=
    \frac{m}{h}\mathbf{I}
    +h\mathbf{K}
    +\mathbf{C}
    +\alpha\mathbf{I}.
  \end{aligned}
  \label{eq:implicit}
\end{equation}
where $\mathbf{I}\in\mathbb{R}^{3N\times3N}$. Partitioning the particles into the prescribed grasp set $\mathcal{C}$ and the free interior, only the free velocities are solved for, with the grasp motion entering through static condensation. Since the linearization is fixed, $\mathbf{A}$ is constant and can be factorized once and reused across sub-steps and rollouts. The implicit formulation is unconditionally stable~\cite{baraff1998}, allowing substantially fewer integration sub-steps than the reference model.

The coarse model retains gravity, elastic deformation, and imposed grasp motion while omitting finer deformation effects. The resulting approximation error grows over the prediction horizon and is corrected by the learned residual. The coarse simulator is run open-loop, maintaining its own $(\mathbf{x}^c,\mathbf{v}^c)$ state throughout each rollout rather than re-synchronizing with the corrected prediction.

\textbf{Observations and latent state.}
During inference, only the grasp points are observed,
\begin{equation}
  {\mathbf{y}}_k
  =
  \left[\mathbf{x}_k^{\mathcal{C}},\mathbf{v}_k^{\mathcal{C}}\right]
  \in\mathbb{R}^{|\mathcal{C}|\times6},
  \label{eq:obs}
\end{equation}
Position inputs are expressed relative to the grasp-point centroid $\mathbf{o}_k=\frac{1}{|\mathcal{C}|}\sum_{i\in\mathcal{C}}\mathbf{x}_{k,i}$, fixed over each prediction horizon, and normalized using training-set statistics. As in~\cite{cRVAE}, a latent variable $\mathbf{z}_k\in\mathbb{R}^{d_z}$ captures the unobserved cloth state. During training it is sampled from the recognition network $q_\phi(\mathbf{z}_k\mid\mathbf{x}_k,\mathbf{h}_k)$, while during inference it is sampled from the conditional prior $p_\psi(\mathbf{z}_k\mid\mathbf{h}_k)$. Both are diagonal Gaussian and use the reparameterization trick, with the prior learned toward the posterior through the KL term in \eqref{eq:elbo}.

\textbf{Residual decoder.}
The decoder is a convolutional network over the $N_u\times N_v$ cloth mesh that predicts a per-particle residual conditioned on the coarse prediction, latent state, recurrent memory, and grasp commands,
\begin{equation}
  \boldsymbol{\Delta}_{k+1}
  =
  D_\theta\!\left(
  \mathbf{x}^{c}_{k+1};
  [\mathbf{z}_k,\mathbf{h}_{k+1},\mathbf{g}_k]
  \right).
  \label{eq:dec}
\end{equation}
The conditioning variables are incorporated through feature-wise linear modulation~\cite{perez2018film}. The corrected prediction is
\begin{equation}
  \hat{\mathbf{x}}_{k+1}
  =
  \mathcal{P}_{\mathbf{g}_k}\!\left(
  \mathbf{x}^{c}_{k+1}
  +\mathbf{r}\odot\boldsymbol{\Delta}_{k+1}
  \right),
  \label{eq:add}
\end{equation}
where $\mathbf{r}\in\mathbb{R}^{3}$ contains the per-axis residual standard deviations, $\odot$ denotes element-wise multiplication, and $\mathcal{P}_{\mathbf{g}_k}$ sets the grasped particles to their commanded positions.

\textbf{Training objective.}
The model is trained over rollouts of length $T$ using
\begin{equation}
  \mathcal{L}
  =
  \sum_{k=1}^{T}
  \left\|
  \left(\hat{\mathbf{x}}_k-\mathbf{x}_k\right)
  \oslash\mathbf{r}
  \right\|^2
  +
  \beta\sum_{d=1}^{d_z}
  \max\!\left(
  \mathrm{KL}_d(q_\phi\|p_\psi),
  \lambda
  \right),
  \label{eq:elbo}
\end{equation}
where $\oslash$ denotes element-wise division. The KL term uses free bits to prevent premature latent collapse. Training uses backpropagation through time and scheduled sampling, gradually reducing the use of measured states in favor of model predictions. Additional training details are given in Sec.~\ref{sec:exp}.

\subsection{Hierarchical Planning}
\label{sec:hplan}
The coupled cloth--robot planning problem requires both global exploration of collision-free motions and local refinement into smooth, executable trajectories. We therefore adopt a hierarchical strategy that combines sampling- and optimization-based methods. MPPI performs task-level exploration using the learned cloth model, while a subsequent MPC layer refines the resulting trajectory subject to the robot's kinematic constraints. Both stages use the proposed reduced-order model of the mobile dual-arm robot, avoiding the high-dimensional full robot model during planning while retaining the relevant base and end-effector motions.

\subsubsection{Full and Reduced-order Model of Robot}
\label{sec:reduced_model}

\begin{figure}[t]
  \centering
  \includegraphics[
    width=.9\columnwidth,
    trim={.5cm 0cm .5cm .5cm},
    clip
  ]{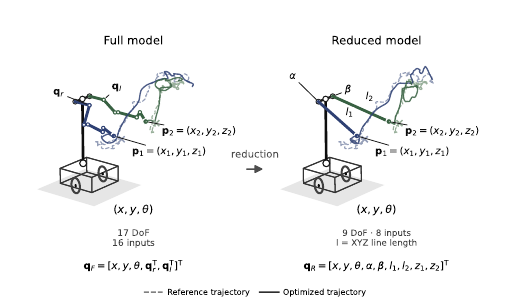}
  \caption{The performance of the MPC refinement layer is evaluated under the full model vs. The reduced-order model as prediction models for the planner. The full model considers the complete kinematic chains of both arms, while the reduced-order model approximates each arm with a simple link that captures only the corresponding end-effector position and not the arms configurations. The reference trajectory $\mathbf{g}^{r}_{k:k+H}$ generated by the MPPI planner contains jerky motion. Both planners track the reference trajectory and provide a smoother reference to the whole-body controller.}
  \label{fig:fulll_red_kinematics_model}
\end{figure}

The full and reduced-order kinematic models of the robot are shown in Fig.~\ref{fig:fulll_red_kinematics_model}. The \emph{full model} retains the base configuration and both seven-joint arms, resulting in a total 17 configuration coordinates as $\mathbf{q}_\mathrm{F}=[x,y,\theta,\mathbf q_r^\top,\mathbf q_l^\top]^\top$ and 16 velocities as $\mathbf{\nu}_{\mathrm F}=[\mathrm v,\mathrm \omega,\dot{\mathbf q}_r^\top,\dot{\mathbf q}_l^\top]^\top$, where $\mathrm v$ and $\mathrm \omega$ denote the linear and angular velocities of the base, respectively. On the other hand, the \emph{reduced-order model} replaces each arm with an equivalent variable-length link, resulting in less configuration coordinates as $\mathbf{q}_\mathrm{R}=[x,y,\theta,\alpha,\beta,l_1,l_2,z_1,z_2]^\top$ and eight velocities, $\mathbf{\nu}_{\mathrm R}=[\mathrm v,\mathrm \omega,\dot{\alpha},\dot{\beta},\dot{l}_1,\dot{l}_2,\dot{z}_1,\dot{z}_2]^\top$. Here, $\alpha$ and $\beta$ denote the link azimuths relative to the base, while $z_i$ denotes the corresponding end-effector height. The lengths $l_i$ represent the spatial shoulder-to-end-effector distances.

The reduced-order model is sufficient for whole-body planning because the planner only needs to determine feasible motions for the nonholonomic mobile base of the robot and its two end-effectors. The detailed arm joint configurations are then handled by the whole-body controller.
The robot workspace is approximated by bounds on the link lengths $l_i$, the azimuth angles $\alpha$ and $\beta$, and the end-effector heights $z_i$, while the nonholonomic constraint of the nonholonomic base is preserved. Thus, the reduced-order model retains the main constraints required for planning while using a much smaller state and input space, which reduces the computational cost of the optimization.

\begin{figure}[t]
  \centering
  \includegraphics[
    width=.7\columnwidth,
    trim={0cm .5cm 0cm 0cm},
    clip
  ]{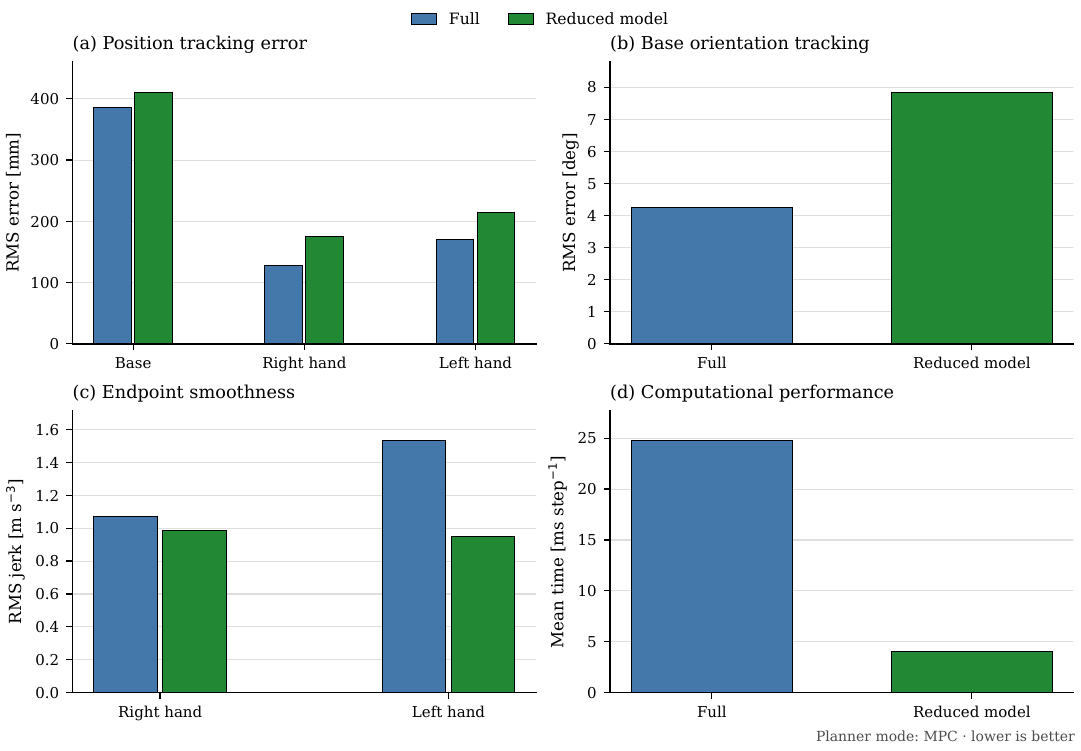}
  \caption{Comparison of the planner performance between full and the reduced-order types. Both versions effectively track the reference $\mathbf{g}_k^r$ while smoothing its jerky motion. However, the optimizer using the full model has a 5x higher computational cost than the one using the reduced model.}
  \label{fig:plot_prediction_model_performance}
\end{figure}

\subsubsection{Sampling-Based Trajectory Generation}
\label{sec:mppi}

The upper planning layer uses model predictive path integral (MPPI) control to explore candidate motions and generate collision-aware reference trajectories for the robot grasp points and mobile base. Sampling avoids the need for analytical gradients of the cost, which depends on the predicted cloth configuration and collision distances evaluated through p-cRVAE. The reduced-order robot model in Sec.~\ref{sec:reduced_model} enables efficient rollout while retaining the relevant base and end-effector motions.

At each control instant, MPPI samples reduced-order velocity sequences around a nominal sequence, 
\begin{equation} 
\boldsymbol{\nu}^{(j)}_{k:k+H-1} = \bar{\boldsymbol{\nu}}_{k:k+H-1} + \boldsymbol{\epsilon}^{(j)}_{k:k+H-1}, 
\end{equation}
where $H$ is the prediction horizon and $\boldsymbol{\epsilon}^{(j)}$ is the exploration noise for rollout $j$. Each sequence is propagated through the reduced-order model and mapped to the robot grasp positions through forward kinematics, $\mathbf{g}^{r}_k=h(\mathbf{q}_{\mathrm R,k})$. Together with the observed human grasp motion, these commands are provided to p-cRVAE to predict the cloth configuration over the horizon. Each rollout is evaluated using
\begin{equation}
    S^{(j)}
    =
    \sum_{k=0}^{H-1}
    \left(
        c^{\mathrm{col},(j)}_k
        +
        c^{\mathrm{shape},(j)}_k
        +
        c^{\mathrm{act},(j)}_k
    \right).
\end{equation}
The nominal sequence is updated using
\begin{equation}
\begin{aligned}
    w^{(j)}
    &=
    \frac{
        \exp\!\left[-(S^{(j)}-S_{\min})/\lambda\right]
    }{
        \sum_m
        \exp\!\left[-(S^{(m)}-S_{\min})/\lambda\right]
    },\\
    \bar{\boldsymbol{\nu}}
    &\leftarrow
    \bar{\boldsymbol{\nu}}
    +
    \sum_j w^{(j)}
    \boldsymbol{\epsilon}^{(j)},
\end{aligned}
\label{eq:mppi_weights}
\end{equation}
where $S_{\min}=\min_j S^{(j)}$ and $\lambda$ is the temperature parameter. 

The MPPI cost combines collision avoidance, preservation of the desired cloth shape, and robot motion,
\begin{equation}
\begin{aligned}
&c^{\mathrm{col}}_k ={}
  w_{\mathrm{obs}}\sum_{i=1}^{N}
  [\delta_{\mathrm{obs}}-d(\hat{\mathbf{x}}_{k,i},\mathcal{O})]_{+}^{2}\\
&+w_{\mathrm{bdy}}\sum_{i=1}^{N}
  [\delta_{\mathrm{bdy}}-d(\hat{\mathbf{x}}_{k,i},\mathcal{B})]_{+}^{2}
+w_{\mathrm{rob}}
  [\delta_{\mathrm{rob}}-d(\mathcal{B},\mathcal{O})]_{+}^{2},
\end{aligned}
\end{equation}
where $w_{\mathrm{obs}}, w_{\mathrm{bdy}}, w_{\mathrm{rob}}, w_{\mathrm{shape}}, w_u,$ and $w_v$ are the respective cost weights, $[a]_+=\max(a,0)$, $d(\cdot,\cdot)$ is the signed distance between two bodies, negative when they overlap, $\mathcal{O}$ denotes the obstacle, and $\mathcal{B}(\mathbf{q}_{\mathrm R})$ denotes the robot body, approximated by two boxes for the driving base and torso. The first two terms penalize the cloth approaching the obstacle and robot body, respectively, while the third penalizes the robot approaching the obstacle. The margins $\delta_{(\cdot)}$ provide a buffer before contact.

The shape cost penalizes deformation relative to the nominal carried configuration,
\begin{equation}
c^{\mathrm{shape}}_k =
  \frac{w_{\mathrm{shape}}}{N}
  \sum_{i=1}^{N}
  \left\|
  \hat{\mathbf{x}}_{k,i}
  -\mathbf{x}^{\mathrm{ref}}_i
  -\boldsymbol{\xi}_k
  \right\|_2,
\end{equation}
where $w_{\mathrm{shape}}$ is the shape-preservation weight and $\boldsymbol{\xi}_k$ denotes the mean displacement of the human-held particles, thereby removing the global translation induced by the human motion.

The action cost is
\begin{equation}
\begin{aligned}
c^{\mathrm{act}}_k ={}&
  w_u\|\boldsymbol{\nu}_k
  \oslash\boldsymbol{\nu}^{\max}\|_2^2
+
  w_v\left[
  \left(\frac{v_k}{v^{\max}}\right)^2+
  \left(\frac{\omega_k}{\omega^{\max}}\right)^2
  \right]\\
&+
  \frac{w_r}{l_b^2}
  \sum_{i=1}^{2}
  [|l_{k,i}-l_{\mathrm{nom}}|-l_b]_{+}^{2},
\end{aligned}
\label{eq:mppi_action_cost}
\end{equation}
where $w_u$ and $w_v$ weight arm and base motion, respectively, $\oslash$ denotes element-wise division, $l_{k,i}$ is the shoulder-to-end-effector distance of arm $i$, and $l_{\mathrm{nom}}\pm l_b$ defines the arm comfort band. The extension penalty implicitly shifts motion from the arms to the mobile base as the arms approach the comfort-band boundary, with $w_r/w_v$ determining the effective transition distance while the band remains below $l_{\max}$ to preserve a nonzero incentive for base motion.

\subsubsection{MPC-Based Trajectory Refinement}
\label{sec:planner}

The trajectories $\mathbf{g}^{r}_{k:k+H}$ generated by the MPPI planner in Sec.~\ref{sec:mppi} lie within the admissible configuration space of the robot. However, these trajectories are noisy, and an MPC layer is further needed to refine them before sending the reference commands to the whole-body controller. The MPC layer can use either the full-order or the reduced-order mode of the robot, as shown in Fig.~\ref{fig:fulll_red_kinematics_model}, as its prediction model in the optimization.
The state vector of the optimizer based on the full-order model is $\mathbf{s}_\mathrm{F}=[\mathbf q_{\mathrm F}^\top,\boldsymbol\nu_\mathrm F^\top]^\top\in\mathbb R^{33}$, while for the reduced-order model it is $\mathbf{s}_\mathrm{R}=[\mathbf q_{\mathrm R}^\top,\boldsymbol\nu_\mathrm R^\top]^\top\in\mathbb R^{17}$. The tracking output is defined as $\mathbf x=h(\mathbf q_{\mathrm{F}/\mathrm{R}})=[x,y,\theta,\mathbf p_1^\top,\mathbf p_2^\top]^\top$, where $\mathbf p_i$ denotes the Cartesian position of the corresponding end-effector, and $h$ denotes the forward kinematics of the respective model. The term $\mathbf u=\dot{\boldsymbol\nu}$ defines the control input vector. Over a prediction horizon of $\bar{\mathrm H}$ steps, the objective is

\begin{equation}
\begin{aligned}
J={}&\|\mathbf e_H\|_{\mathbf P}^{2}
+\sum_{k=0}^{\bar{\mathrm H}-1}\Big(
\|\mathbf e_k\|_{\mathbf Q}^{2}
+\|\boldsymbol\nu_k-\boldsymbol\nu_k^{\mathrm{ref}}\|_{\mathbf R}^{2}\\
&\hspace{20mm}
+\|\mathbf u_k\|_{\mathbf S}^{2}
+\|\boldsymbol\phi_k\|_{\mathbf W}^{2}\Big),
\end{aligned}
\end{equation}
where $\mathbf e_k=h(\mathbf q_{\mathrm{F/R},k})-\mathbf x_k^{\mathrm{ref}}$, $\boldsymbol\phi_k=[\alpha_k,\beta_k]^\top$, and $|\mathbf a|_{\mathbf M}^{2}=\mathbf a^\top\mathbf M\mathbf a$ denotes the weighted quadratic cost of a generic vector $\mathbf a$. These terms penalize the tracking error, deviation from the reference velocity, acceleration effort, and link-azimuth deviation.
The optimization enforces the discretized kinematics, configuration, velocity, and acceleration bounds, together with $\|\dot{\mathbf p}_i\|_2\leq0.18\,\mathrm{m\,s^{-1}}$. The problem is solved using acados SQP-RTI with explicit Runge--Kutta integration, $\bar{\mathrm H}=20$, and $\Delta t=0.1\,\mathrm s$. The one-step-ahead prediction supplies the next base and end-effector commands.

The performance of both frameworks is compared in Fig.~\ref{fig:plot_prediction_model_performance}. Both show comparable performance in smooth tracking of the reference trajectory $\mathbf{g}^{r}_{k:k+H}$ generated by the MPPI planner, with the tracking performance of the optimization using the full-order model being marginally higher, while the jerk RMS of the optimization using the reduced-order model is marginally lower. However, the biggest advantage of using the reduced-order formulation is that it achieves comparable tracking performance with approximately 80\% lower computation time than the full-order model.

\subsection{Robot Whole Body Control}
\label{sec:wbc}
For the purpose of tracking the refined reference command by the end-effectors of the robot, the hierarchical whole-body control framework in \cite{forouhar2025whole} is employed. At the first task priority, the stacked positions of the two end-effectors, $\mathbf{x}_{e}=[\mathbf{x}_{r}^{\top},\mathbf{x}_{l}^{\top}]^{\top}\in\mathbb{R}^{6}$, define the primary task. At the second task priority, the position and orientation $\mathbf{x}_{b}=[x, y,\theta]^{\top}\in\mathbb{R}^{3}$ of the mobile base define the secondary task. The corresponding task Jacobians $\mathbf{J}_{e}$ and $\mathbf{J}_{b}$ map the reduced-order whole-body pseudo-velocity $\boldsymbol{\nu}$ to the respective task velocities, $\dot{\mathbf{x}}_{e}=\mathbf{J}_{e}\boldsymbol{\nu}$ and $\dot{\mathbf{x}}_{b}=\mathbf{J}_{b}\boldsymbol{\nu}$, where $\boldsymbol{\nu}$ spans the admissible subset of the generalized whole-body velocity $\dot{\mathbf{q}}_{\mathrm{w}}$ satisfying the base's non-holonomic constraint. The remaining system redundancy is resolved in the null-space to accomplish lower-priority objectives, such as maintaining the arm configurations as close as possible to their default posture.

\section{Experiments}
\label{sec:exp}

\subsection{Training of p-cRVAE}
\label{sec:training}

\begin{figure*}[!htb]
    \centering

    \begin{subfigure}[t]{0.245\textwidth}
        \centering
        \vspace{0pt}
        \includegraphics[width=\linewidth]{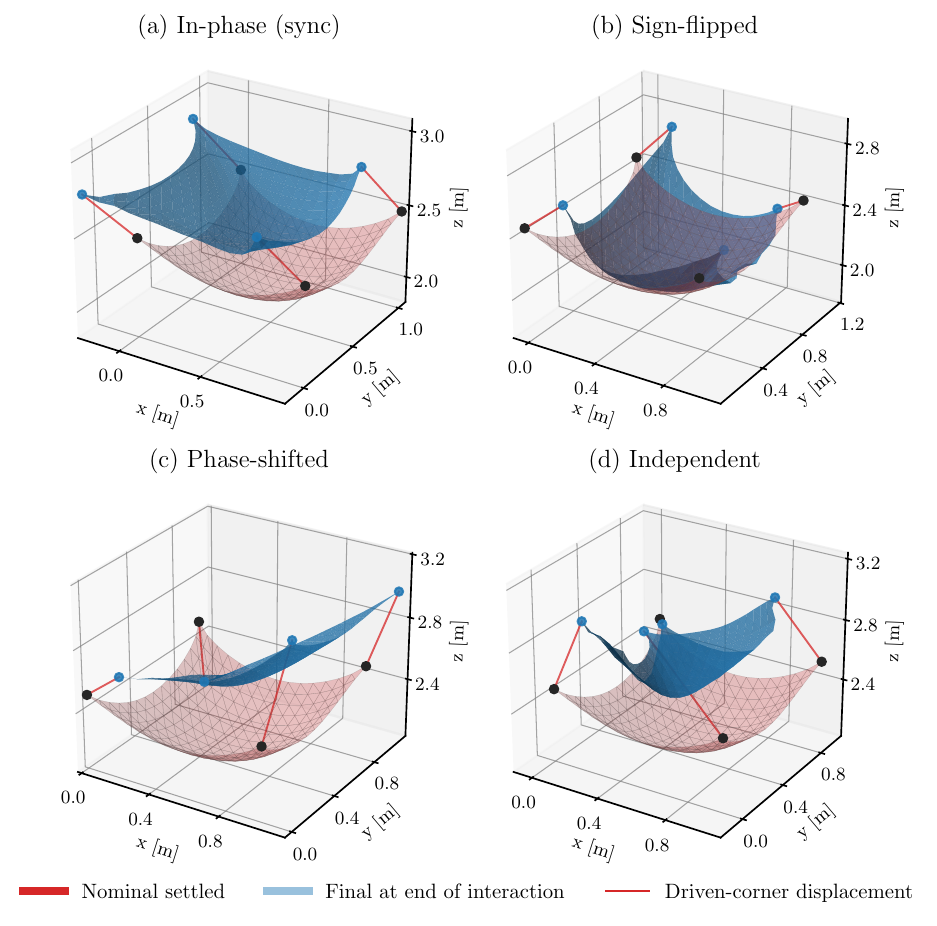}
        \caption{}
        \label{fig:exp_trainingdataset}
    \end{subfigure}%
    \hspace{0.02\textwidth}%
    \begin{subfigure}[t]{0.660\textwidth}
        \centering
        \vspace{0pt}
        \includegraphics[width=\linewidth]{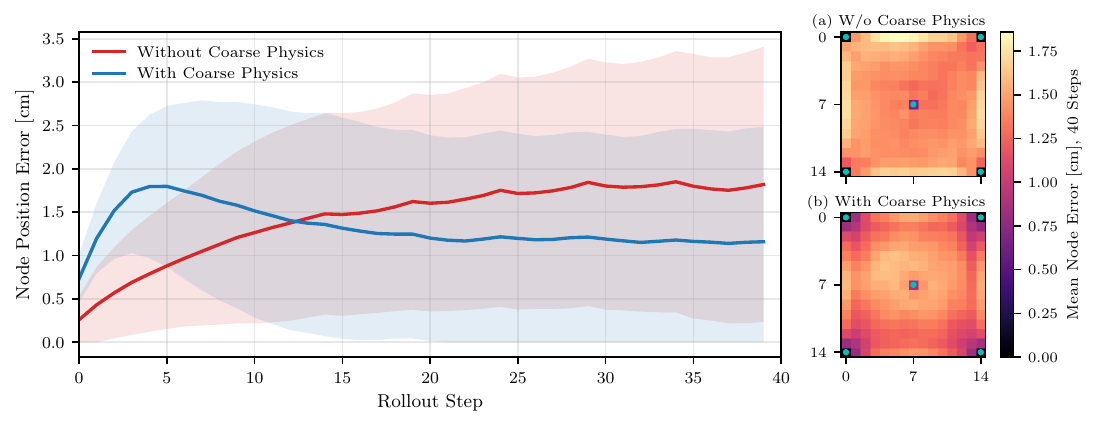}
        \caption{}
        \label{fig:exp_predictionerror}
    \end{subfigure}

    \caption{Training data and open-loop prediction accuracy.
    (\subref{fig:exp_trainingdataset}) Grasp-point coordination patterns used to
    generate the dataset, shown as the settled and final sheet for a
    representative episode of each pattern.
    (\subref{fig:exp_predictionerror}) Per-node prediction error over the rollout,
    with and without the coarse backbone (mean $\pm$ one standard deviation over
    held-out episodes), and spatial error maps over the mesh on a shared scale.
    Teal markers denote the measured points.}
    \label{fig:exp_training}
\end{figure*}

The p-cRVAE is trained on $930$ episodes generated using the mass-spring model of Sec.~\ref{sec:cloth}, each $60\,\mathrm{s}$ long with $\Delta t=0.1\,\mathrm{s}$. Grasp points are driven kinematically to decouple the data from a specific robot and cover configurations beyond the reduced-order robot's reachable set. For every non-empty subset of the four corners, motions follow one of three patterns: \emph{sign}, with shared motion and per-corner sign changes; \emph{phase}, with shared motion and phase offsets; or \emph{independent}, with separately driven corners. This produces both coordinated and highly sheared configurations (Fig.~\ref{fig:exp_training}\subref{fig:exp_trainingdataset}), while slow grasp motions maintain the quasi-static regime.

We train with Adam for $300$ epochs using a batch size of $32$ and an initial learning rate of $10^{-3}$ with cosine decay, holding out $15\%$ of episodes for validation. The rollout length increases from $12$ to $40$ steps, with scheduled sampling reducing the grounded fraction from $1.0$ to $0.3$. The KL weight is annealed to $\beta=0.3$ with a free-bits threshold of $\lambda=0.05$ nats per latent dimension. The latent and memory dimensions are $d_z=16$ and $d_h=256$, respectively, and the coarse physics model uses four sub-steps per control period.

Figure~\ref{fig:exp_training}\subref{fig:exp_predictionerror} evaluates open-loop prediction on held-out episodes with periodic measurement injection. We compare p-cRVAE with the same architecture without the coarse physics backbone. Without the backbone, error increases throughout the rollout, whereas p-cRVAE reaches an early peak and converges to a lower plateau. Despite its initial approximation error, the physics backbone improves long-horizon prediction, with spatial error maps showing reduced error near the measured grasp points and residual error concentrated in the less-constrained cloth interior.

\subsection{Evaluation in simulation}
\label{sec:closedloop}

\begin{figure}[!t]
    \centering

    \subfloat[MuJoCo simulation.]{
        \includegraphics[width=.69\linewidth]
        {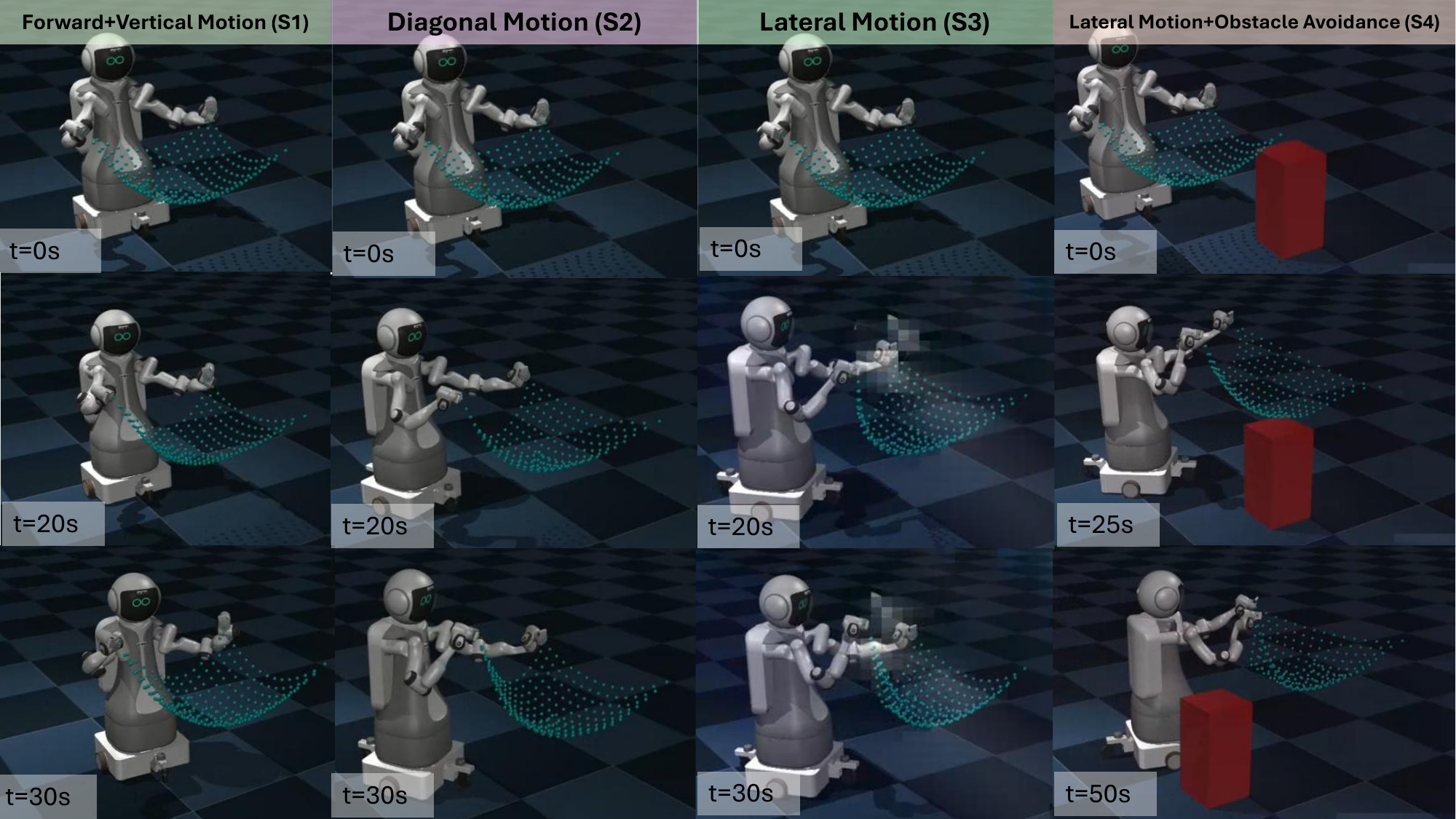}
        \label{fig:mujoco_sim}
    }

    \par\vspace{2mm}

    \subfloat[Real-robot experiment. ]{
        \includegraphics[width=.69\linewidth]
        {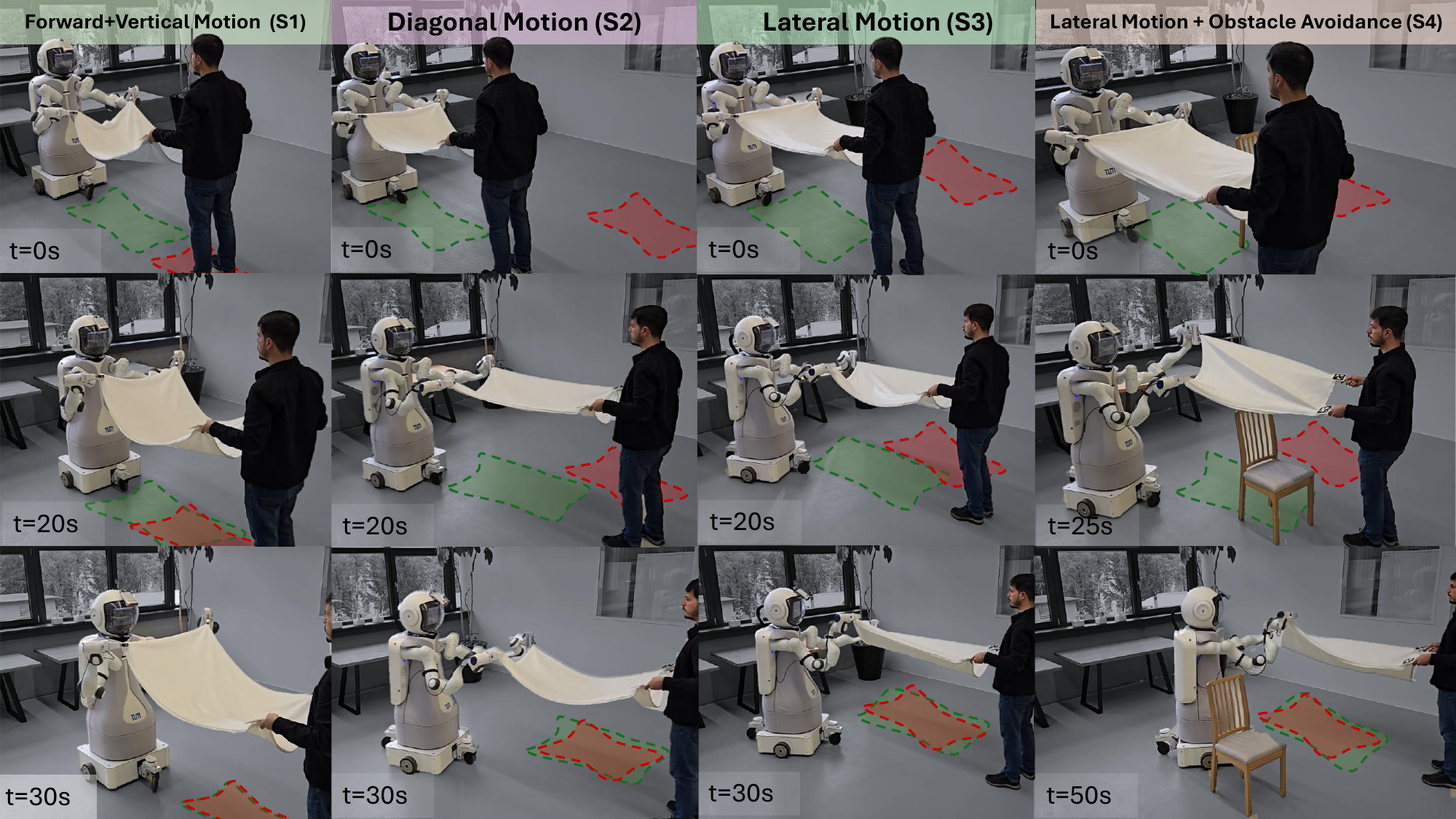}
        \label{fig:real_exp}
    }

    \par\vspace{1mm}

    \caption{Comparison of the MuJoCo simulations and corresponding real-robot experiments for the four human--robot co-transportation scenarios. In real experiment, the green shadow represents the current position of the cloth and red shadow shows the goal position.}
    \label{fig:exp_mujoco}
\end{figure}

We evaluate the complete framework in closed-loop MuJoCo simulation (Fig.~\ref{fig:mujoco_sim}), using the mass-spring model of Sec.~\ref{sec:cloth} as the cloth plant and prescribing the human grasp points along a straight-line trajectory. The MPPI planner, whole-body motion planner, and controller run as separate processes using the same interfaces as the physical system; only the plant is simulated.

\textbf{Scenarios.}
We consider four increasingly challenging scenarios at slow and fast carry speeds within a fixed time budget (see Fig.~\ref{fig:exp_mujoco}(a)). Scene~1 (S1) requires primarily forward and vertical motion, Scene~2 (S2) adds diagonal motion subject to the non-holonomic constraint, Scene~3 (S3) requires lateral motion and base reorientation, and Scene~4 (S4) adds an obstacle to the lateral carry. Clearance is evaluated only in S4.

\textbf{Baselines.}
We consider two baselines isolating the contributions of p-cRVAE and whole-body motion planning. Baseline~I replaces the predicted-shape cost with corner following and removes the cloth--obstacle cost while retaining robot collision avoidance, following the corner-following formulation in~\cite{sirintuna2024}. Baseline~II retains the proposed MPPI planner and cost but bypasses the whole-body motion planner, passing its reference directly to the whole-body controller.

\textbf{Results.}
Table~\ref{tab:main} reports accumulated and final cloth deformation, control effort, and minimum cloth--obstacle clearance for Scene~4. Baseline~I performs comparably in the simplest scene, where corner following approximates the desired cloth motion, but deformation increases for lateral and shearing motions and the method fails in Scene~4 at both speeds by driving the cloth into the obstacle. This highlights the importance of predicting the unactuated cloth interior for collision avoidance.

Baseline~II produces larger deformation because the sampled references are not reconciled with the full robot's kinematic and non-holonomic constraints. Its apparently favorable clearance in Scene~4 results from large tracking errors that prevent the cloth from reaching the intended trajectory. The proposed method instead maintains low deformation across all scenes and speeds, with control effort comparable to Baseline~I and positive obstacle clearance.

\textbf{Ablations.}
Table~\ref{tab:ablation} evaluates Scene~4. A shorter prediction horizon provides insufficient reaction time, reducing obstacle clearance, while longer horizons offer limited benefit at increased computational cost. Increasing the obstacle weight improves clearance at the expense of shape tracking, whereas increasing the shape weight improves tracking but can cause collision. The selected weights provide a balance between the two objectives.

% generated by paper/table_latex.py -- do not edit by hand
\begin{table*}[t]
  \centering
  \setlength{\tabcolsep}{3.5pt}
  \footnotesize
  \caption{End-to-end results on the four evaluation scenes at two carry speeds. Mean$\pm$std over three repeats of a fixed 300-frame run. $\sum c_{\mathrm{anchor}}$ and $\sum c_{act}$ are integrated over the run. dev$_T$ is the midpoint-anchored mean node deviation of the full 225-node sheet at the final frame. Scenes 1--3 carry no obstacle, so clearance is undefined there. B-I: corner-following only, no shape prediction. B-II: no whole-body NMPC.}
  \label{tab:main}
  \resizebox{0.95\textwidth}{!}{%
  \begin{tabular}{lrrrrrrrrrrrr}
    \toprule
\textbf{Scene}, $v_h$ & \multicolumn{3}{c}{$\sum c_{\mathrm{anc}}$ [$10^6$] $\downarrow$} & \multicolumn{3}{c}{dev$_T$ [cm] $\downarrow$} & \multicolumn{3}{c}{$\sum c_{\mathrm{act}}$ [$10^3$] $\downarrow$} & \multicolumn{3}{c}{clr [cm] $\uparrow$} \\
    \cmidrule(lr){2-4} \cmidrule(lr){5-7} \cmidrule(lr){8-10} \cmidrule(lr){11-13}
     & B-I & B-II & Prop. & B-I & B-II & Prop. & B-I & B-II & Prop. & B-I & B-II & Prop. \\
    \midrule
    S1, 0.075 & 0.10{\tiny$\pm$0.02} & 1.65{\tiny$\pm$0.03} & 0.13{\tiny$\pm$0.01} & 32.5{\tiny$\pm$1.4} & 50.0{\tiny$\pm$0.2} & 28.7{\tiny$\pm$4.2} & 2.00{\tiny$\pm$0.05} & 2.07{\tiny$\pm$0.14} & 2.03{\tiny$\pm$0.02} & -- & -- & -- \\
    S1, 0.125 & 0.14{\tiny$\pm$0.01} & 1.70{\tiny$\pm$0.01} & 0.16{\tiny$\pm$0.00} & 33.1{\tiny$\pm$0.5} & 49.8{\tiny$\pm$0.3} & 32.2{\tiny$\pm$0.3} & 2.06{\tiny$\pm$0.05} & 2.03{\tiny$\pm$0.04} & 2.02{\tiny$\pm$0.03} & -- & -- & -- \\
    \addlinespace[2pt]
    S2, 0.075 & 0.27{\tiny$\pm$0.01} & 1.93{\tiny$\pm$0.02} & 0.24{\tiny$\pm$0.01} & 29.5{\tiny$\pm$2.0} & 67.6{\tiny$\pm$1.9} & 29.5{\tiny$\pm$2.3} & 2.06{\tiny$\pm$0.02} & 2.50{\tiny$\pm$0.12} & 2.11{\tiny$\pm$0.17} & -- & -- & -- \\
    S2, 0.125 & 0.53{\tiny$\pm$0.02} & 2.03{\tiny$\pm$0.03} & 0.29{\tiny$\pm$0.01} & 26.1{\tiny$\pm$1.8} & 66.7{\tiny$\pm$0.3} & 29.3{\tiny$\pm$0.8} & 2.03{\tiny$\pm$0.04} & 2.39{\tiny$\pm$0.11} & 2.11{\tiny$\pm$0.05} & -- & -- & -- \\
    \addlinespace[2pt]
    S3, 0.075 & 0.39{\tiny$\pm$0.01} & 1.23{\tiny$\pm$0.01} & 0.25{\tiny$\pm$0.01} & 22.3{\tiny$\pm$0.3} & 44.6{\tiny$\pm$0.5} & 30.1{\tiny$\pm$1.2} & 2.24{\tiny$\pm$0.03} & 2.45{\tiny$\pm$0.09} & 2.26{\tiny$\pm$0.07} & -- & -- & -- \\
    S3, 0.125 & 0.42{\tiny$\pm$0.01} & 1.30{\tiny$\pm$0.03} & 0.27{\tiny$\pm$0.00} & 24.6{\tiny$\pm$5.4} & 44.7{\tiny$\pm$0.1} & 27.7{\tiny$\pm$0.2} & 2.19{\tiny$\pm$0.06} & 2.57{\tiny$\pm$0.06} & 2.31{\tiny$\pm$0.04} & -- & -- & -- \\
    \addlinespace[2pt]
    S4, 0.075 & 0.76{\tiny$\pm$0.02} & 2.12{\tiny$\pm$0.03} & 0.44{\tiny$\pm$0.01} & 25.1{\tiny$\pm$1.3} & 93.3{\tiny$\pm$0.5} & 27.6{\tiny$\pm$1.7} & 2.13{\tiny$\pm$0.04} & 2.42{\tiny$\pm$0.05} & 2.17{\tiny$\pm$0.06} & $-$13.0{\tiny$\pm$0.4} & 6.2{\tiny$\pm$0.2} & 2.1{\tiny$\pm$0.5} \\
    S4, 0.125 & 1.01{\tiny$\pm$0.01} & 2.34{\tiny$\pm$0.05} & 0.53{\tiny$\pm$0.01} & 26.3{\tiny$\pm$1.2} & 92.9{\tiny$\pm$0.1} & 28.3{\tiny$\pm$1.9} & 2.09{\tiny$\pm$0.04} & 2.41{\tiny$\pm$0.08} & 2.18{\tiny$\pm$0.04} & $-$11.2{\tiny$\pm$0.9} & 7.0{\tiny$\pm$0.1} & 1.3{\tiny$\pm$0.3} \\
    \bottomrule
  \end{tabular}}
\end{table*}

\begin{table}[t]
  \centering
  \setlength{\tabcolsep}{3.5pt}
  \footnotesize
  \caption{Ablations on scene4 yobs40 at 0.125\,m/s, the cell with the tightest time budget. Mean$\pm$std over three repeats. $H{=}40$ is the proposed horizon and is read from the same runs as \emph{Prop.} in Table~\ref{tab:main}.}
  \label{tab:ablation}
  \resizebox{\columnwidth}{!}{%
  \begin{tabular}{lrrrrr}
    \toprule
\textbf{Ablation} & $\sum c_{\mathrm{anc}}$ [$10^6$] $\downarrow$ & dev$_T$ [cm] $\downarrow$ & $\sum c_{\mathrm{act}}$ [$10^3$] $\downarrow$ & clr [cm] $\uparrow$ & solve [ms] $\downarrow$ \\
    \midrule
    $H{=}10$ & 0.51{\tiny$\pm$0.00} & 29.4{\tiny$\pm$0.9} & 1.11{\tiny$\pm$0.03} & $-$3.7{\tiny$\pm$0.4} & 31.9{\tiny$\pm$0.2} \\
    $H{=}40$ (ours) & 0.53{\tiny$\pm$0.01} & 28.3{\tiny$\pm$1.9} & 2.18{\tiny$\pm$0.04} & 1.3{\tiny$\pm$0.3} & 96.4{\tiny$\pm$0.3} \\
    $H{=}60$ & 0.65{\tiny$\pm$0.04} & 25.3{\tiny$\pm$4.4} & 2.66{\tiny$\pm$0.09} & 1.6{\tiny$\pm$0.4} & 139.3{\tiny$\pm$0.6} \\
    $w_{\mathrm{obs}}\!\times\!10$ & 0.66{\tiny$\pm$0.02} & 25.0{\tiny$\pm$2.3} & 2.15{\tiny$\pm$0.02} & 9.7{\tiny$\pm$0.2} & 96.6{\tiny$\pm$1.3} \\
    $w_{\mathrm{shape}}\!\times\!10$ & 5.92{\tiny$\pm$0.30} & 22.6{\tiny$\pm$3.0} & 2.13{\tiny$\pm$0.06} & $-$7.3{\tiny$\pm$0.6} & 97.1{\tiny$\pm$0.3} \\
    \bottomrule
  \end{tabular}}
\end{table}

\subsection{Real-robot validation}
The four scenarios introduced in Sec.~\ref{sec:closedloop} are further evaluated in real experiments (see Fig.~\ref{fig:exp_mujoco}(b)). In the real setup, the whole-body motion planner and controller operate in closed loop, whereas the high-level co-transportation trajectories generated by the MPPI planner are replayed from simulation and therefore executed open loop. As shown in Fig.~\ref{fig:real_exp}, the robot successfully coordinates with the human to transport the blanket in all four scenarios, exhibiting behavior consistent with the simulation. In the obstacle-avoidance scenario S4, the robot deforms and lifts the blanket over the obstacle while simultaneously coordinating its two arms and non-holonomic mobile base to keep the end-effectors within the feasible arm workspace and satisfy the base non-holonomic constraint. Despite the high-level planner being executed open loop, these experiments demonstrate that the reference trajectories generated by MPPI using the p-cRVAE cloth model and the reduced-order robot model are feasible for the real robot--cloth system. 

\section{Discussion \& Conclusion}
\label{sec:discussion}
This work introduced a framework for human--robot co-transportation of a deformable object using a dual-arm nonholonomic mobile robot. Closed-loop simulations show that the upper-layer MPPI planner combines the physics-residual cloth model (p-cRVAE) and reduced-order robot model to generate trajectories that preserve the reference cloth configuration while satisfying robot constraints. The lower-layer MPC refines these trajectories into smooth and feasible references for whole-body control. While performance is comparable to the baselines in obstacle-free scenarios, the proposed framework provides improved obstacle avoidance. The sim-to-real gap of the mass-spring model has been studied in~\cite{cRVAE}, and corner-point perception provides a relatively simple observation interface. However, closing the remaining sim-to-real gap and achieving reliable closed-loop operation on the physical system remain important directions for future work. We will also investigate learning a family of deformable-object models and incorporating latent-state filtering to improve robustness to model uncertainty and partial observations.

\section{Acknowledgement}
We acknowledge funding from StMWi Bayern through the Lighthouse Initiative Geriatronics (LongLeif GaPa gGmbH, Project Y), the European Union’s Horizon Europe programme through FlexCycle (Grant No. 101189600), and BMFTR through the ``Souver\"{a}n. Digital. Vernetzt'' programme, joint project 6G-life (Project No. 16KIS2414). We also acknowledge NVIDIA Corporation for their support to Tampere University for this work through the Academic Grant Program.

\small{\bibliography{main}}
\bibliographystyle{IEEEtran}

\end{document}